# AI-Care: A Conversational Agentic System for Task Coordination in Alzheimer’s Disease Care

Preyash Yadav
*Department of Computer Science*
*University of California, Davis*
Davis, USA
preyadav@ucdavis.edu

Michelle Cohn
*Department of Linguistics*
*University of California, Davis*
Davis, USA
mdcohn@ucdavis.edu

Priyanka Koppolu
*Department of Neurology*
*University of California, Davis*
Sacramento, USA
prkoppolu@health.ucdavis.edu

Hritvik Agarwal
*Department of Neurology*
*University of California, Davis*
Davis, USA
hritvik.agarwal@gmail.com

Amey Gohil
*Department of Computer Science*
*University of California, Davis*
Davis, USA
agohil@ucdavis.edu

Tejas Patil
*Department of Computer Science*
*University of California, Davis*
Davis, USA
tpatil@ucdavis.edu

Sasha Pimento
*Department of Neurology*
*University of California, Davis*
Sacramento, USA
spimentol@ucdavis.edu

Alyssa Weakley
*Department of Neurology*
*University of California, Davis*
Sacramento, USA
aweakley@health.ucdavis.edu

**Abstract — Individuals with Alzheimer’s disease (AD) and Alzheimer’s disease-related dementia (ADRD) experience memory and thinking changes that impact their ability to use digital daily management tools. For example, adding an event to a digital calendar requires multiple steps that may act as barriers to independent use for individuals with AD/ADRD. This paper presents AI-Care, a conversational agentic artificial intelligence (AI) layer built on top of a remote caregiving platform co-designed with people with AD/ADRD. AI-Care is designed to reduce the cognitive load on individuals with AD/ADRD when managing everyday tasks such as setting calendar reminders and organizing to-do lists through natural-language interaction with a voice-first chatbot. The system uses a LangGraph-based stateful orchestration approach in which each request passes through sanitization, intent classification, context loading, safety checks, deterministic slot collection, tool execution, and response composition. Safety-critical responses, particularly around medications and allergies, are grounded in caregiver-verified records rather than free-form model generation. The system does not make autonomous medical or treatment decisions. Incomplete or ambiguous requests are handled through controlled multi-turn clarification rather than silent failure or guessing. The system supports both typed and spoken input, with voice output through ElevenLabs text-to-speech. Longer responses are chunked before synthesis to avoid rushed playback. A preliminary pilot with four individuals with mild-to-moderate AD/ADRD showed that users found the system trustworthy, competent, and likable, and were able to complete the evaluated coordination tasks through conversation. We describe the design goals, system architecture, safety controls, and findings from this formative evaluation.**

## Introduction

Millions of individuals with Alzheimer’s disease (AD) and Alzheimer’s disease-related dementia (ADRD) experience cognitive decline that interferes with their ability to complete daily tasks. Routine tasks including calendar management, setting reminders, and tracking to-do lists become harder to sustain independently. This is concerning because missed actions, such as taking medications, can result in serious health consequences that threaten their ability to age-in-place, a desire strongly held by 75-90% of older adults [1]. This strong desire leads at least 25% of older adults with AD/ADRD to live alone [2]. To help their cognitively impaired family members age-in-place, adult children often take on remote caregiving roles and spend much of their time coordinating their parents' daily schedules [3]. As such, remote caregivers desire digital management tools that individuals with AD/ADRD can use independently while allowing caregivers to play a supportive rather than directive role.

Digital caregiving platforms and smart-home systems have improved remote coordination, monitoring, and AD/ADRD engagement in their own care [4]–[7]. However, many such systems rely on menu-driven workflows that require multiple steps for simple actions. For users with cognitive impairment, this interaction overhead can itself become a barrier in daily use. Prior work [8] has observed that individuals with AD/ADRD have a tendency to talk out loud and ask questions as they navigate through tasks. This led our team to hypothesize whether an agentic AI chatbot would aid in task completion and be viewed as a useful and desirable feature.

This paper presents an agentic AI layer (AI-Care) built on top of Interactive-Care (I-Care), a remote caregiving platform designed for older adults with cognitive impairment [5]. AI-Care is designed to reduce cognitive load for care receivers while keeping safety-critical responses grounded in verified records. The system is intentionally scoped to coordination and reminder support, not diagnosis or treatment/care decisions. Here, we report a prototype implementation and preliminary formative findings from a task-based pilot. Our contributions are: (1) a safety-constrained conversational architecture for care coordination grounded in caregiver-verified records, (2) an integration design over I-Care's existing coordination tools and data, and (3) usability and acceptability ratings from end-user interactions.

## Background & Related Work

I-Care is a web-based remote caregiving platform designed for collaborative use by individuals with AD/ADRD and their remote caregivers [5]. It includes a co-designed calendar with step-by-step guided workflows, customizable to-do lists, messaging, and a home page with personal photos and upcoming events [8]. A plug-and-play desktop infrastructure deploys I-Care as a persistent, kiosk-locked system in participants' homes, requiring no login or navigation and supporting remote monitoring and automated recovery [9]. These efforts established an accessible platform and a viable deployment model. However, all interaction within I-Care currently relies on manual, menu-driven workflows. Although individuals with AD/ADRD are able to use and are satisfied with I-Care's iteratively co-designed UI/UX [8], it remains a menu-driven design requiring multiple clicks for even simple task completion. For users with progressive cognitive decline this interaction overhead remains a barrier. AI-Care, the conversational layer described in this paper, was developed to address this challenge.

A growing body of work has explored technologies to support older adults in maintaining independence. Smart-home sensing systems have been used to monitor activities of daily living and detect cognitive decline [4], [10], [11], while web-based caregiving platforms and assistive applications provide structured support for memory and task management [5]–[7]. Companion robots and embodied agents have also been investigated for improving medication adherence, cognitive engagement, and quality of life [12], [13]. Earlier systems often relied on predefined workflows or constrained interaction paradigms, while more recent work explores LLM-based approaches that enable flexible, conversational interaction, Often with fewer task constraints or less explicit safety grounding. These systems demonstrate the potential of technology-assisted care while underscoring the need for reliable, safety-aware designs in real-world settings.

Voice-based interfaces have emerged as a promising modality for accessibility in this population. Prior work shows that older adults often prefer voice interaction over traditional text input [14], and that voice assistants can help reduce loneliness and improve social connectedness for individuals with dementia [15]. Recent work has begun to explore the role of generative and agentic AI in dementia care. Krueger [16] conceptualizes AI-enabled environments through the extended mind thesis, suggesting that ambient AI systems can act as cognitive and affective extensions that support autonomy. Grammenos et al. [17] argue for agentic systems capable of integrating multimodal clinical data for Alzheimer's disease management, while Khalil et al. [18] survey large language model-based agents for elderly care, highlighting opportunities for personalized support alongside concerns around hallucinations, privacy, and accessibility. Finally, Bazgir et al. [19] propose AgenticAD, a multi-agent architecture spanning care planning, retrieval, and analysis tasks, demonstrating coordination across specialized agents but without deployment or user-centered evaluation.

Despite these advances, existing systems either focus on clinical decision-making, broad assistive capabilities, or conceptual architectures, and do not address the constraints of everyday task-related interaction for individuals with cognitive impairment. In particular, to our knowledge, no prior work describes a safety-constrained conversational agent integrated into a deployed caregiving platform, designed specifically for AD/ADRD users, and grounded in caregiver-verified records.

# Methodology

## *Design Goals & Scope*

The agentic AI component, AI-Care, is designed to make daily coordination easier for care receivers without adding cognitive load, as they manage routine tasks like reminders, to-do items, and calendar entries. Rather than navigating menu-driven workflows, users can interact through short back-and-forth conversation, which is easier to follow and less cognitively demanding. The system supports these interactions through multiple input modes, including typed text, voice input via speech-to-text, and optional spoken replies through text-to-speech, making it accessible to older adults with varying comfort levels with technology.

Safety-critical responses, particularly those involving medications and allergies, are grounded in caregiver-verified truth records rather than free-form model generation. The system does not make autonomous medical or treatment decisions; its role is limited to coordination and reminders. When a request is incomplete or unclear, the system asks follow-up questions rather than failing silently or guessing. By design, the assistant prioritizes reliability over broad capability.

To support caregiver oversight, the system confirms completed actions in plain language so the user is not left uncertain about what happened. All state-changing operations are recorded in an audit trail, giving caregivers a way to review what the system has done on the patient's behalf. This also allows for caregiver-in-the-loop feedback to the agentic AI system and modifications to completed actions, if needed. As shown in Figure 1, AI-Care is integrated on top of I-Care.

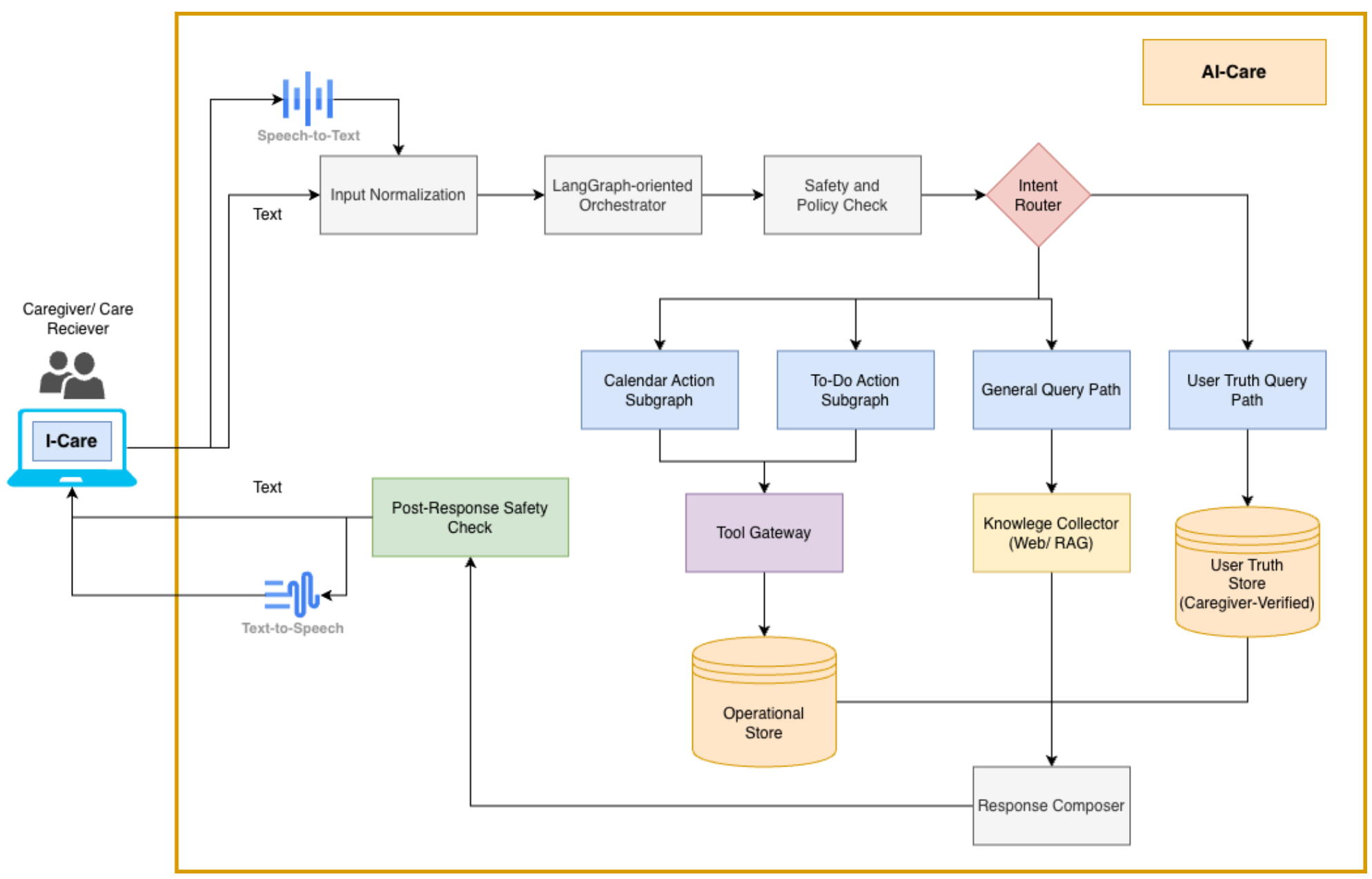

*Figure 1: High-level System Design: AI-Care integrated with I-Care platform*

## *LangGraph-Oriented System Architecture and Runtime Flow*

The system is built as a LangGraph state machine in which each user request is tracked through a typed shared state. This state carries request metadata (session, caregiver, and care receiver identifiers), the raw input and normalized text, classified intent and extracted entities, user context and truth data, pending action state for multi-turn flows, domain-specific execution plans and results, safety flags, and the final response. At a high level, the system is organized into three layers: an interaction layer handling chat and voice, an orchestration layer containing the graph nodes, router, and subgraphs, and a data layer comprising a read-only user truth store, a mutable operational store, and an audit and context store.

When input enters the graph, a sanitization node strips unsafe content such as script injections and null bytes, and normalizes whitespace. A coarse NLU node then classifies the request domain and action type but does not extract slots; slot extraction is handled separately by the schema engine during subgraph execution. A context node loads the caregiver-receiver linkage, truth records, list configuration, medications, allergies, short-term interaction context, and any pending action state from prior turns. Before any action is planned, a pre-safety guardrails node checks allergy and medication policies and verifies permissions.

A router node dispatches the request based on a fixed priority order: explicit cancellations are handled first, then truth queries, then pre-guardrail rejections, then pending schema resumptions, then NLU-confident routes, and finally a general fallback. Truth and general queries are routed directly to their respective response handlers. To-do and calendar requests enter symmetric subgraphs that each follow a fixed sequence of nodes: plan the tool call, validate arguments, check policies, execute the tool, and verify the result.

Within the to-do and calendar subgraphs, a schema engine handles deterministic slot collection. When the subgraph planner selects a candidate action such as “todo.create" or “calendar.modify," a resolver maps it to the specific operation (for example, “calendar.modify" becomes “calendar.update" or “calendar.delete" based on keyword analysis). The schema engine then parses slots from the user’s input in priority order, merges them into the existing partial state using defined rules (for example, a filled slot receiving a different value triggers a conflict), validates them individually and across slots, runs slot-level safety checks, and resolves record-match slots against existing data. If all required slots are ready, the engine builds a typed payload and passes it to the tool executor. If not, it returns a follow-up question targeting the next unresolved required slot. Disambiguation is handled similarly: if a record-match slot finds multiple candidates, the engine returns a numbered list and waits for the user to select one before proceeding.

The system’s tool execution is designed to be deterministic rather than relying on open-ended model generation. The agent’s capabilities are defined through structured configuration files: an agent definition specifies the assistant’s role, scope, and behavioral constraints, while a tools definition registers the specific operations available to it, such as creating a to-do item or updating a calendar event. Each tool is backed by a small, self-contained script that performs one specific task through defined API calls and database operations, following documented I-Care procedures. This keeps individual operations simple, testable, and predictable. If a request does not match any defined tool or domain path, a catch-all handler manages the error gracefully rather than allowing the system to fail silently or produce an ungrounded response.

Multi-turn state is persisted through a dedicated persistence layer that stores one partial action per caregiver-receiver pair. This allows clarification and disambiguation flows to survive across API calls and restarts. When a user replies to a follow-up, the input re-enters the graph at the normalization node with the preserved partial state, forming a controlled multi-turn loop. If an unrelated request occurs mid-flow, the active task is paused and its structured state is stored on a stack. After handling the new request, the system resumes the prior task through explicit clarification prompts, preserving progress without inferring missing information.

The response composer builds the final output strictly from verified data and tool results. A post-response safety node provides a fallback check before delivery. A persist-and-audit node then stores context updates, pending state transitions, safety rejection events, and logs for all write operations. The Runtime flow is depicted in Figure 2.

On the voice input side, spoken input is transcribed using OpenAI’s Whisper ASR system before entering the normalization node. For output, responses are returned as text by default. When voice output is enabled, the text is sent to ElevenLabs for synthesis using a natural-sounding voice profile (Ava US English voice). Longer responses are chunked into smaller segments before being sent to text-to-speech, as testing showed that sending longer text at once resulted in rushed playback that was harder for older adult users to follow.

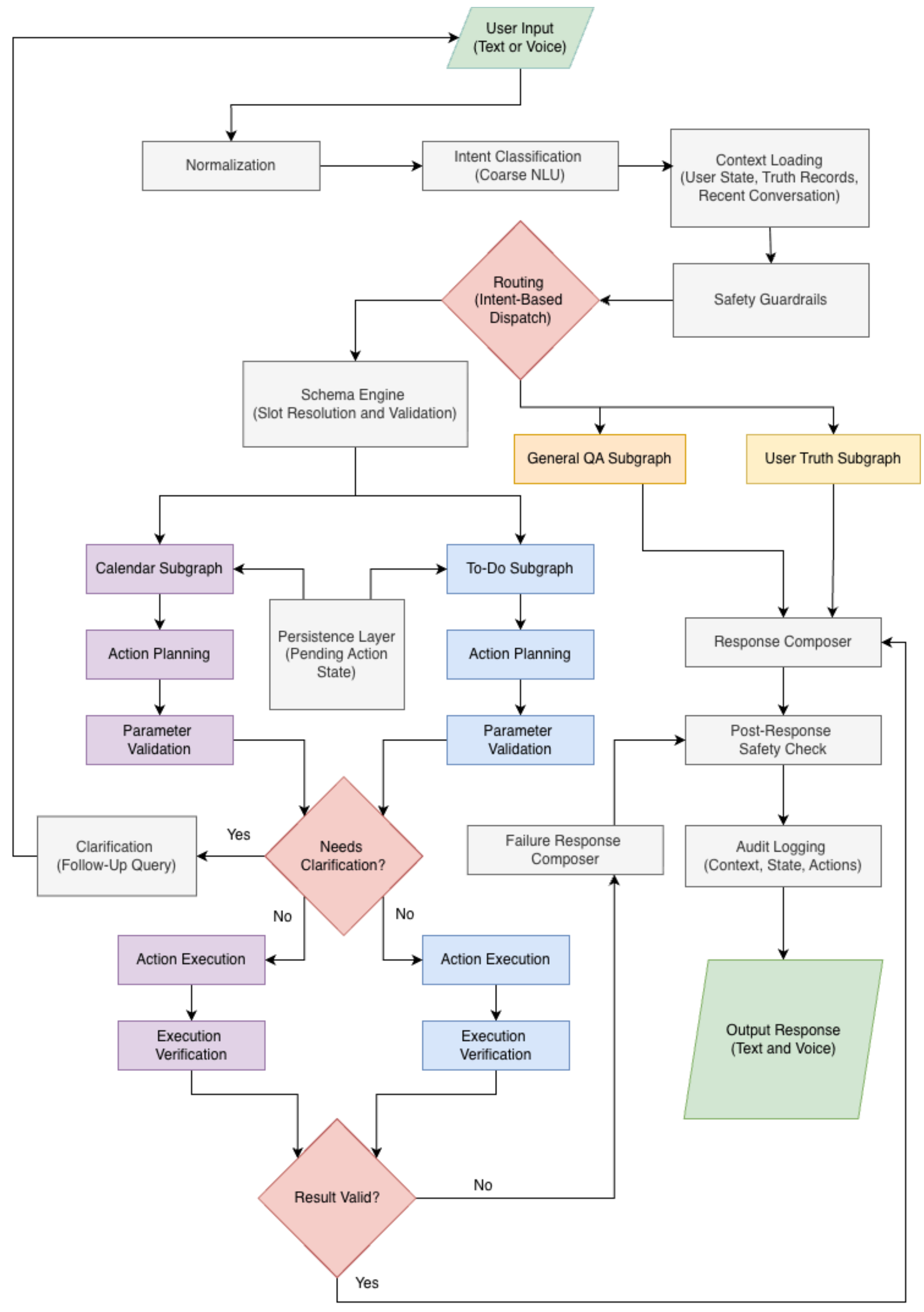

*Figure 2: AI-Care Runtime Flow*

### *Safety, Human Oversight, and Data Governance*

Safety is enforced through three layers. A pre-safety guardrails node runs before any action is planned, checking for allergy conflicts, medication policy violations, out-of-scope requests, and permission issues at the conversational level. Within the schema engine, slot-level safety checks can reject individual slot values before a tool call is ever assembled, for example blocking a to-do item that conflicts with a known allergen even if the overall intent appeared safe. At execution time, a final policy gate validates the complete action payload against authorization and domain rules immediately before any database mutation occurs.

The data layer is split by trust level. A caregiver-governed truth store holds safety-critical facts such as allergies and medications and is treated as read-only from the agent and patient interaction path, with caregiver-governed update workflows. A separate operational store holds editable items such as to-dos, calendar entries, pending action state, and short-term interaction context. This separation prevents routine automation from silently changing high-risk records.

Human oversight is maintained through clarification and disambiguation flows whenever requests are incomplete or ambiguous. The system asks follow-up questions rather than guessing, presents numbered options when multiple record matches exist, and surfaces conflicts when a new slot value contradicts a previously confirmed one. Authorization checks ensure actions are only performed for valid caregiver-receiver links, and identity fields are bound server-side for every tool call. All state-changing operations and safety rejections are audit-logged for traceability. Credentials for model and voice services remain server-side. The current system is a research prototype with active safety controls but is not presented as a fully validated clinical platform at this time.

## Evaluation & Preliminary Findings

Pilot testing was conducted with four individuals with mild-to-moderate AD/ADRD (age M=84 years, SD=4.24; education M=15.5 years, SD=2.52; 50% non-Hispanic white, 25% Hispanic White; 25% Asian; 50% bilingual (English-Spanish; English-Mandarin); 100% female). General familiarity with technology ranged from very low to moderate and they had been interacting with I-Care for 4 weeks prior to evaluation. Participants primarily lived alone (1 user lived with her spouse but received remote care from her daughter) and in the general community (one participant resided in assisted living). Pilot evaluation consisted of interactions with AI-Care external to the I-Care system, where participants were asked to create a to-do list item, add something to their calendar, and check their allergy list.

After the interaction, users were asked to rate their impression of AI-Care on the following dimensions: untrustworthy–trustworthy, machine-like–human-like, incompetent–competent, and unlikeable–likable (5-point Likert scale: Extremely X, Somewhat X, Neither X nor Y, Somewhat Y, and Extremely Y; coded numerically from 1–5) (from (Bartneck et al. 2009; Cohn, Pushkarna, et al. 2024)). Finally, participants were asked, “Does the chatbot seem like a real person? Explain why or why not.” (Cohn, Barreda, et al. 2024) and if they would like to see AI-Care integrated into their I-Care platform.

Ratings for each dimension are provided in Figure 3. As seen, users rated the system as somewhat trustworthy (mean = 4±0.82), competent (mean = 4.5±1.41), and likable (mean = 4.5±0.58). AI-Care was seen as neither human-like nor machine-like (mean = 3±0.58). Qualitative responses for whether AI-Care seemed like a real person varied: 2 users referenced the voice characteristics (“voice sounded natural”, “sometimes, because of the accent”), 1 user brought up top-down beliefs (“no, I can see that it’s a computer and not a real person.”), and one user indicated they were not sure (“in between”). All users indicated perceived usefulness, expressing that they would like to see it integrated into I-Care.

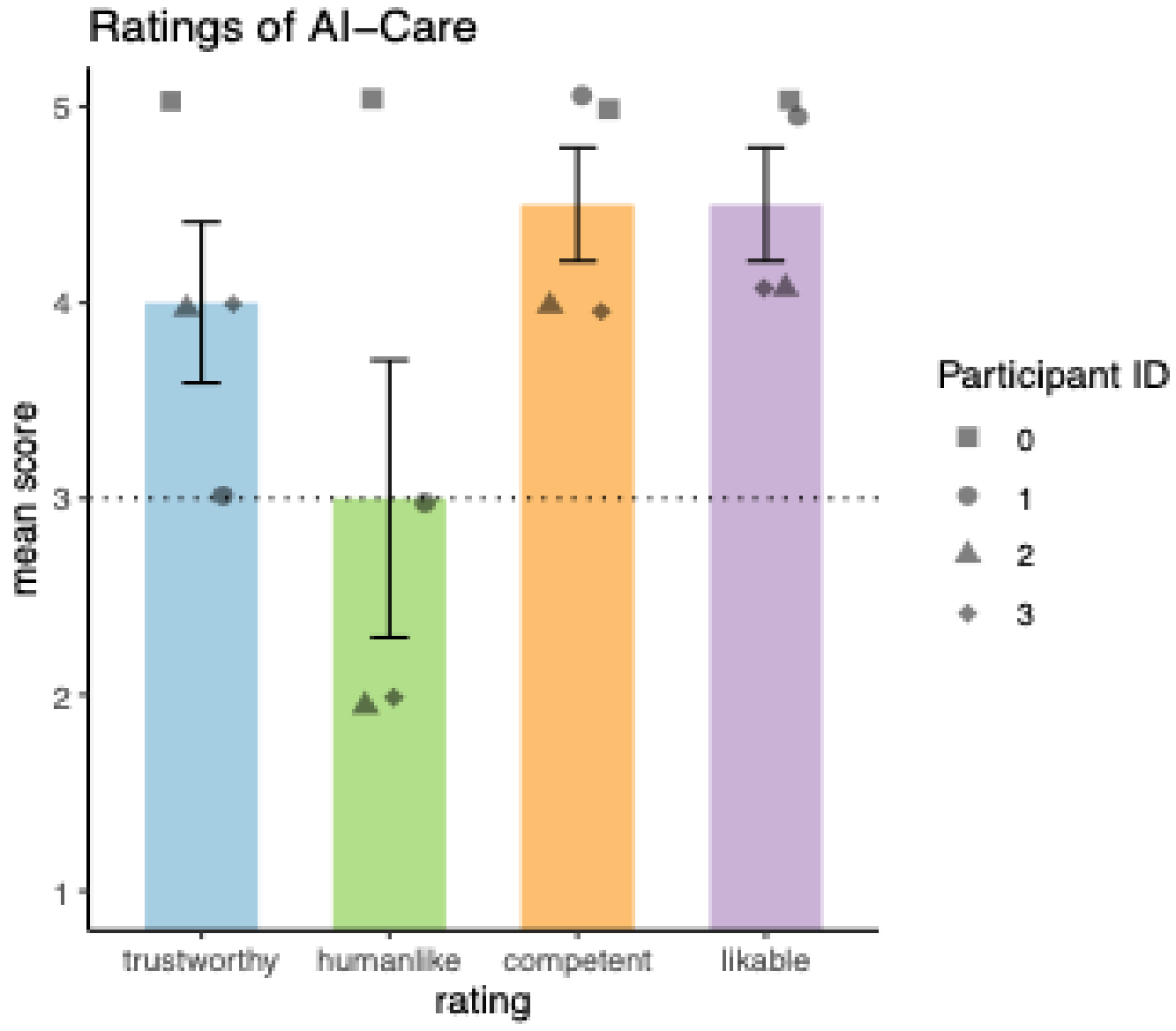

*Figure 3: Participants' mean ratings of AI-Care system. Error bars show standard error of the mean. Points show individual participants' ratings. Values above 3 indicate a positive score.*

## Discussion

This paper presents the system architecture and safety controls of AI-Care, a conversational agentic AI layer built on top of the remote caregiving platform I-Care. Functionally, users were able to interact with AI-Care and complete coordination tasks through conversation, providing initial evidence that the approach is viable for older adults with AD/ADRD. Users perceived AI-Care positively, rating it as trustworthy, competent, and likable. While they did not view it as strongly human-like, perceptions varied: for one user, the voice contributed to a sense of realism, while another's awareness that it was a computer reduced anthropomorphism. This suggests that voice quality supports engagement but is not sufficient on its own, and that individual differences shape user experience. Importantly, all users expressed interest in incorporating AI-Care into their I-Care platform, providing preliminary evidence of perceived usefulness and satisfaction.

## Limitations

There are several limitations of the present pilot study. Most notable is the small sample size that was homogeneous in some respects (e.g., all female) and heterogeneous in others (e.g., bilingual and monolingual speakers). For testing purposes, AI-Care was tested as a standalone component rather than within the full I-Care platform and tasks were structured rather than open-ended, limiting insight into real-world use. AI-Care is also limited to task specific support which may be confusing to individuals with cognitive impairment who have experience with other voice assistant systems.

Finally, while not directly observed in the current study, we acknowledge that voice interfaces can introduce challenges, including short response windows (Addlesee and Eshghi 2024) and automatic speech recognition (ASR) errors for older adults and individuals with cognitive impairment (Cohn et al. 2026; Werner et al. 2019). In our current prototype, we avoid strict response timeouts, allowing users extended time to respond. In the future, we will implement adaptive turn-taking or structured response windows, to improve human-computer interaction and accessibility.

## Future Work

As the present findings are exploratory, a larger and more diverse study with full platform integration and longer-term use is needed to assess real-world performance and impact. A key next step is conducting parallel task-based evaluations in which users perform the same coordination tasks (e.g., creating a calendar event) using both the existing menu-driven workflow and the conversational interface, measuring task completion time, error rates, assistance needed, and user preferences to identify where each approach is most effective. These findings will inform how both interaction modes can coexist and guide refinement of the conversational system, including improvements to AI-Care's workflow such as more adaptive interaction pacing, enhanced clarification strategies, and more robust handling of multi-turn and interrupted requests based on observed user behavior.

# Conclusion

This paper presented AI-Care, a safety-constrained conversational agentic AI system designed to reduce cognitive load for individuals with AD/ADRD managing everyday coordination tasks. Our pilot study suggests that a constrained, safety-focused conversational approach is a promising direction for improving accessibility in caregiving platforms.